\documentclass[11pt,a4paper]{article}
\usepackage{sty}
\usepackage{times}
\usepackage{latexsym}

\usepackage{tikz}
\usepackage{tikz-dependency}
\usepackage{amsmath}

\title{Unsupervised Part-of-Speech Induction}
\author{Omid Kashefi\\
Intelligent Systems Program\\
University of Pittsburgh\\
\texttt{kashefi@cs.pitt.edu}\\}
\date{}

\begin{document}

\maketitle

\begin{abstract}
\noindent
Part-of-Speech (POS) tagging is an old and fundamental task in natural language processing. While supervised POS taggers have shown promising accuracy, it is not always feasible to use supervised methods due to lack of labeled data. In this project, we attempt to unsurprisingly induce POS tags by iteratively looking for a recurring pattern of words through a hierarchical agglomerative clustering process. Our approach shows promising results when compared to the tagging results of the state-of-the-art unsupervised POS taggers. \\
\end{abstract}

\section{Introduction}
Part-of-Speech (POS) is the morphosyntactic category of words such as \emph{Noun}, \emph{Verb} or \emph{Adjective}. As words are the first principle of each natural language, so determining POS category of words is one of the most fundamental tasks in natural language processing (NLP).

As each word can belong to more than one POS category, the POS tagging task is harder than having just a simple dictionary of words with their POS category and is actually disambiguation from morphosyntactic categories of words. As an example, the word ``duck'' could be a \emph{Noun} or also a \emph{Verb} for example in sentence ``You can duck to hide.''

Old and still state of the art in automatic POS tagging is using supervised machine learning algorithms to learn hand-crafted POS features in large text corpora. The first English computational-friendly corpus with POS features was Brown corpus \cite{francis1964brown} with about 1 million tagged words. Nowadays, there are lots of annotated corpora available for NLP and computational linguistic research such as WSJ corpus \cite{charniak2000bllip} and Penn Treebank \cite{marcus1993building}.

Multiple machine learning methods are experienced to model POS features such as support vector machine \cite{gimenez2004svmtool}, hidden Markov model \cite{brants2000tnt}, maximum entropy \cite{manning2011part}, maximum entropy Markov models \cite{denis2009coupling} and condition random field \cite{sun2014structure}. The current reported state of the art per token POS tagging accuracy is 97.55\%\footnote{See \texttt{http://aclweb.org/aclwiki/index.php?title=POS\_Tagging\_(State\_of\_the\_art)}}.

Although supervised learning approaches towards POS tagging show promising accuracy, many resource-poor languages lack suitably annotated corpora to benefit from these methods. Moreover, hand-crafting features are expensive and time-consuming so it motivates research on unsupervised approaches toward feature induction in NLP. Likewise, unsupervised POS induction is an active area of research in NLP.

\section{Unsupervised POS Induction}
From the unsupervised perspective, we look at POS tagging as a clustering problem wherein we will try to assign words into different clusters that later we will name them as words' morphosyntactic or POS classes. 

As the grammar or syntax of a language is formed from the sequence of word categories (i.e. POS tags), we believe words with the same context are most likely to belong to the same word category. Therefore, we iteratively look for words with similar context (in this work we consider tri-grams as context) and put them into the same cluster.

\subsection{Clustering Approach}
In order to implement a multi-iteration clustering approach, we chose to use hierarchical agglomerative clustering (HAC) that is an iterative clustering approach by nature. 

We start from each word belonging to a distinct cluster as shown in Figure~\ref{fig:hac0} for a corpus containing two sentences \texttt{``I fed a black dog''} and \texttt{``I saw the black dog''}. Then we look for words $w_x$ that are most likely to belong to the same category. We find $w_x$ such that they occurred in a tri-gram context with lowest probability distribution as given in Equation \ref{eq:wx}. 

\begin{equation} \label{eq:wx}
    w_x = \arg \min H(P(w_{x}|w_{i},w_{i+1}))
\end{equation}

We chose $w_x$ with lowest entropy in tri-grams probability distribution because, consider a big-gram $w_{i},w_{i+1}$ that can be followed by 200 words in the corpus and another $w_{j},w_{j+1}$ that can be followed by only 5 words. Those 5 words are most likely to belong to the same category than those 200 words as $w_{j},w_{j+1}$ context is more specific than $w_{j},w_{j+1}$ that can be followed by any categories of words.

\begin{figure}
    \centering
    
    \begin{tikzpicture}[->,>=stealth',shorten >=0pt,auto,node distance=.4cm,main node/.style={draw, text height=.8em, rectangle}]
        \node[main node] (1) {I};
        \node[main node] (2) [right=of 1]{fed};
        \node[main node] (3) [right=of 2]{a};
        \node[main node] (4) [right=of 3]{black};
        \node[main node] (5) [right=of 4]{dog};
        \node[->,>=stealth',shorten >=0pt,auto,node distance=.4cm] (6) [right=of 5]{...};
        \node[main node] (7) [right=of 6]{I};
        \node[main node] (8) [right=of 7]{saw};
        \node[main node] (9) [right=of 8]{the};
        \node[main node] (10) [right=of 9]{black};
        \node[main node] (11) [right=of 10]{dog};
        
        \node[main node] (111) [below=of 3, yshift=-2cm]{X};
        
        \path[thick]
        (3) edge node [auto] {} (111)
        (9) edge node [auto] {} (111);
        
    \end{tikzpicture}
    \caption{POS Induction Clustering Process, Iteration 1 \label{fig:hac1}}
\end{figure}
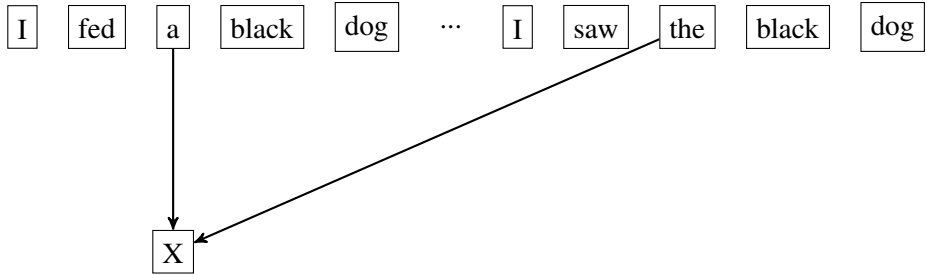

After choosing $w_x$ on each iteration, we put them into the cluster $c_x$ as shown in Figure \ref{fig:hac1}. In the next iterations, $w_x$ can belong to a new cluster or one of the existing clusters. Moreover, as the syntax is a product of a sequence of word categories, so choosing $c_x$ for $w_x$ also depends on the neighbor's category. Therefore, we extend the model such that choosing $w_x$ and $c_x$ maximize the model's probability as given in Equation \ref{eq:model}.

\begin{multline} \label{eq:model}
    (w_x, c_x) = \arg \max_{w_x, c_x} \prod P(w_x|c_x)P(c_x|c_{i}c_{i+1}) \\
    = \arg \max_{w_x, c_x} \sum \log P(w_x|c_x)P(c_x|c_{i}c_{i+1})
\end{multline}

Therefore, after we build $w_x$ candidates based on the entropy of their tri-gram probability distribution, we start to assigning them into either new or existing cluster $c_x$ and chose the $(w_x, c_x)$ pair such that it maximizes the model's probability. 

Looking closet to Equation \ref{eq:model}, we are trying to find the best cluster sequence $c_x$ that can generate the observed words sequence $w_x$. So our model is actually an HMM and instead of trying all possible emission and transition pairs, we use Baum–Welch (a.k.a forward-backward) algorithm \cite{baum1970maximization} to efficiently estimate them on each iteration.

\begin{figure}
    \centering

    \begin{tikzpicture}[->,>=stealth',shorten >=0pt,auto,node distance=.4cm,main node/.style={draw, text height=.8em, rectangle}]
      \node[main node] (1) {I};
      \node[main node] (2) [right=of 1]{fed};
      \node[main node] (3) [right=of 2]{a};
      \node[main node] (4) [right=of 3]{black};
      \node[main node] (5) [right=of 4]{dog};
      \node[->,>=stealth',shorten >=0pt,auto,node distance=.4cm] (6) [right=of 5]{...};
      \node[main node] (7) [right=of 6]{I};
      \node[main node] (8) [right=of 7]{saw};
      \node[main node] (9) [right=of 8]{the};
      \node[main node] (10) [right=of 9]{black};
      \node[main node] (11) [right=of 10]{dog};

    \end{tikzpicture}
    \caption{POS Induction Clustering Process, Initial State \label{fig:hac0}}
\end{figure}
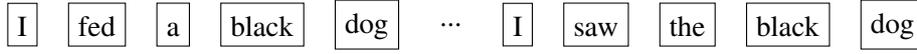

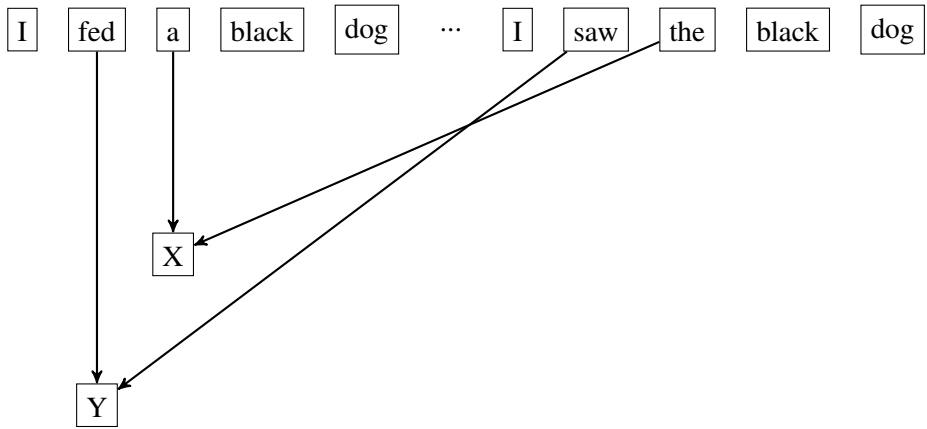
\begin{figure}
    \centering
    
    \begin{tikzpicture}[->,>=stealth',shorten >=0pt,auto,node distance=.4cm,main node/.style={draw, text height=.8em, rectangle}]
        \node[main node] (1) {I};
        \node[main node] (2) [right=of 1]{fed};
        \node[main node] (3) [right=of 2]{a};
        \node[main node] (4) [right=of 3]{black};
        \node[main node] (5) [right=of 4]{dog};
        \node[->,>=stealth',shorten >=0pt,auto,node distance=.4cm] (6) [right=of 5]{...};
        \node[main node] (7) [right=of 6]{I};
        \node[main node] (8) [right=of 7]{saw};
        \node[main node] (9) [right=of 8]{the};
        \node[main node] (10) [right=of 9]{black};
        \node[main node] (11) [right=of 10]{dog};
        
        \node[main node] (111) [below=of 3, yshift=-2cm]{X};
        
        \node[main node] (211) [below=of 2, yshift=-4cm]{Y};
        
        \path[thick]
        (3) edge node [auto] {} (111)
        (9) edge node [auto] {} (111)
        (2) edge node [auto] {} (211)
        (8) edge node [auto] {} (211);
        
    \end{tikzpicture}
    \caption{POS Induction Clustering Process, Iteration 2 \label{fig:hac2}}
\end{figure}

We continue the process to next iterations and as the context now has less variability (as we merged some words into their cluster), we have more chance to find other candidates as shown in Figure \ref{fig:hac2}.

HAC allow us to continue the clustering process until we have any desired process, so we iterate until we cluster the text into 13 categories. We chose this number of clusters to be compatible with the recommended POS tag set of EAGLES morphosyntactic annotation standard \cite{leech1996eagles} as it is more focused on major POS tags (i.e. Noun and Verb) comparing to other standards like Penn TreeBank \cite{marcus1993building} that have finer grain POS tags and can be more representative of word categories.

\section{Evaluation}
We trained our model on Penn TreeBank that contains about 1 million words from WSJ \cite{charniak2000bllip} with POS and treebank annotation. We chose to use this corpus because it is annotated with POS tag and we can easily compare our inducted POS tags with its manually annotated tags as the gold standard.

The accuracy of our model on inducing correct POS tags of Penn TreeBank is 68\% as shown in Table \ref{tab:acc} under PTB column.  It must be noted that we collapse the Penn TreeBank's POS tags into their major categories as directed in EAGLES to be compatible with our induced POS tag set.

\begin{table}
    \centering
    \begin{tabular}{*{8}{c}}
      \hline
      \hline
       &  \multicolumn{5}{c}{\textbf{Unsupervised}} &~~ & \textbf{Supervised}\\
       \cline{2-6} \cline{8-8}
       &  \multicolumn{2}{c}{\small \textbf{Our Approach}} & \small \textbf{Clark} & \small \textbf{Berg} & \small \textbf{Goldwater} &~~ & \small \textbf{TnT}\\
       \cline{2-3}
       &  \small \textbf{PTB} & \small \textbf{Brown} &  &  &  & & \\
    
        \multicolumn{8}{c}{~}\\
    
      \textbf{Accuracy} & 68\% & 73\% & 71\% & 75\% & 76\% &~~ & 86\%\\
    
      \hline
    \end{tabular}
    \caption{POS Tagging Accuracy}
    \label{tab:acc}
\end{table}

In order to have a better understanding of how good our unsupervised method works, we need to evaluate it such that can be compared with state-of-the-art unsupervised POS taggers. As the common way of evaluating both supervised and unsupervised POS taggers are computing their accuracy on detecting correct POS tags of an out-of-domain data, we trained a second-order HMM tagger on Penn Treebank annotated with our induced POS tags and evaluated it on tagging Brown corpus \cite{francis1964brown} as an out-of-domain test set. 

The accuracy of the HMM tagger trained with our corpus is depicted in Table \ref{tab:acc}. We also include the accuracy results of tagging Brown corpus for \newcite{clark2003combining}, \newcite{berg2010painless}, and \newcite{goldwater2007fully} that are unsupervised approaches towards POS induction, and unknown word tagging accuracy for TnT \cite{brants2000tnt} that is the best HMM tagger of-the-shelf. 

As it is shown, the accuracy of our POS tagger is comparable with state-of-the-art unsupervised POS taggers and we believe if we consider more features such as morphological similarity of the words we can beat other existing unsupervised taggers.

\section{Related Work}
As the POS is a simple word-level feature of natural languages and can be learned easily with machine learning approaches, almost every new approaches towards NLP are first examined with POS tagging problem. Although unsupervised approaches towards NLP is in research market for a while, the amount of works on POS induction is remarkably understudied comparing to other new trends and we have a few works focusing on unsupervised POS induction. Notable works in POS induction are as follow:

\newcite{brown1992class} uses bigrams as features and a greedy agglomerative hierarchical clustering algorithm. They tried to optimize the probability of corpus based on the probability of the word belonging to a latent class and probability of the latent class of the previous word.

\newcite{clark2003combining} used a similar model as \cite{brown1992class} but adds another level of clustering to take word types (first round clusters) into account for the next iteration. They also considered the morphological similarity of words as a feature.

\newcite{biemann2006unsupervised} uses a graph clustering algorithm called Chinese Whispers that is based on contextual similarity. It first clusters the most frequent 10,000 words based on their context then consider the number of shared neighbors between two words in a 4-word context window.

\newcite{goldwater2007fully} is based on a standard HMM for POS tagging using bigram model as the feature. They placed a Dirichlet priors over the multinomial parameters defining the state-state and state-emission distributions and uses a collapsed Gibbs sampler to infer the hidden tags.

\newcite{berg2010painless} considers character trigrams and capitalization as features and uses HMM model, but assumes that the state-state and state-emission distributions are logistic.

\newcite{yatbaz2012learning} considers vector paradigmatic representations of words along with morphological and orthographic similarities as features and used a modified k-means clustering to determine syntactic categories. It means they extend the presentation of a word with other related words form an ontological database and used this new extended presentation to find similar words in the context. However, we doubt that this work could be considered fully unsupervised as it uses ontological features.

All of these works tried to induce POS tags from raw text but the way we chose the words that are most likely to have the same POS tag by considering words with the lowest entropy in probability distribution of its context is novel and it might be even more efficient than other techniques as the model we employed is by far easier but our performance is just slightly lower than them (we even beat one of the unsupervised taggers).

\section{Conclusion}
In this project, we worked on unsupervised POS induction. We build our model relying on the fact that grammar of a language or an observed sequence of words are the product of the sequence of their categories so we tried to find the best category sequence that can generate the observed text. We iteratively used this model to cluster words, initiated with candidates that occurred in context with the lowest impurity.

Our unsupervised approach shows comparable results against state-of-the-art unsupervised POS taggers on tagging out-of-domain data. We believe employing a model with more features, such as the morphological similarity between words, we can beat state-of-the-art unsupervised taggers.

Finally, given the ever-increasing amounts of digitized data, we believe that unsupervised methods should be reconsidered and we believe that someday unsupervised tools may perform as well as costly supervised tools and bring new horizons to NLP research.

\bibliographystyle{acl}

\end{document}